%% file: Main.tex
\documentclass[letterpaper, 10 pt, conference]{ieeeconf}  

\input{Format}

\begin{document}
\input{Title}
\input{Abstract}
\input{Introduction}

\input{Related-Work}
\input{GripperDesign}
\input{SimulationPlatform}
\input{Conclusion}
\bibliographystyle{IEEEtran}
\bibliography{ref}
\end{document}

%% file: Format.tex
\usepackage{amsmath,amsfonts}
\usepackage{algorithmic}
\usepackage{algorithm}
\usepackage{array}
\usepackage[caption=false,font=normalsize,labelfont=sf,textfont=sf]{subfig}
\usepackage{textcomp}
\usepackage{stfloats}
\usepackage{url}
\usepackage{verbatim}
\usepackage{graphicx}
\usepackage{cite}
\usepackage{xcolor}
\usepackage{xspace}

\makeatletter
\let\NAT@parse\undefined
\makeatother
\usepackage[hidelinks]{hyperref}
\hypersetup{
   linkcolor=blue,
   breaklinks=true,   
   colorlinks=true,   
   citecolor=blue,
   urlcolor=blue
}

\newcommand \figref{Figure~\ref}
\newcommand \grippername{DeGrip\xspace}

\IEEEoverridecommandlockouts  

%% file: Title.tex
\title{\textbf{\grippername}: A Compact Cable-driven Robotic \textbf{Grip}per for Desktop \textbf{D}isass\textbf{e}mbly}

\author{Bihao Zhang$^{1,\dagger}$, Davood Soleymanzadeh$^{1,\dagger}$, Xiao Liang$^{2,*}$, and Minghui Zheng$^{1,*}$
\thanks{$^{1}$Bihao Zhang, Davood Soleymanzadeh, and Minghui Zheng are with the J. Mike Walker '66 Department of Mechanical Engineering, Texas A\&M University, College Station, TX, USA (\tt\footnotesize e-mail: bhzhang@tamu.edu; davoodso@tamu.edu; mhzheng@tamu.edu).}
\thanks{$^{2}$Xiao Liang is with the Zachry Department of Civil and Environmental
Engineering, Texas A\&M University, College Station, TX 77843 USA (\tt\footnotesize e-mail:
xliang@tamu.edu).}
\thanks{$\dagger$: equal contribution; $*$: corresponding authors}
\thanks{This work was supported by the USA National Science Foundation under Grants No. 2422826 and No. 2422640.}}

\maketitle

%% file: Abstract.tex
\begin{abstract}
Intelligent robotic disassembly of end-of-life (EOL) products has been a long-standing challenge in robotics. While machine learning techniques have shown promise, the lack of specialized hardware limits their application in real-world scenarios. We introduce \grippername, a customized gripper designed for the disassembly of EOL computer desktops. \grippername provides three degrees of freedom (DOF), enabling arbitrary configurations within the disassembly environment when mounted on a robotic manipulator. It employs a cable-driven transmission mechanism that reduces its overall size and enables operation in confined spaces. The wrist is designed to decouple the actuation of wrist and jaw joints. We also developed an EOL desktop disassembly environment in Isaac Sim to evaluate the effectiveness of \grippername. The tasks were designed to demonstrate its ability to operate in confined spaces and disassemble components in arbitrary configurations. The evaluation results confirm the capability of \grippername for EOL desktop disassembly.
\end{abstract}

%% file: Introduction.tex
\section{Introduction} \label{sec:intro}
Disassembly of end-of-life (EOL) products is essential for sustainability and advancing the circular economy. The current global demand for natural resources far exceeds what the planet can sustainably provide. Therefore, innovative solutions are necessary for efficient resource management, and increasing recycling rates \cite{sliwowski2025reassemble}. Intelligent robotic systems have the potential to play a significant role in this effort \cite{meng2022intelligent}.

Industrial robotic manipulators are widely used for assembly tasks in manufacturing, where they are manually programmed to perform repetitive operations. In contrast, robotic disassembly of EOL products is rarely realized at an industrial scale. Disassembly poses unique challenges: products from different manufacturers exhibit a large variation in size and configuration, and their physical condition is often uncertain. Therefore, the robotic system needs to move beyond the pre-programmed behaviors and autonomously adapt to the given tasks.

Recent advances in imitation learning (IL) \cite{chi2023diffusion} and reinforcement learning (RL) \cite{noseworthy2025forge} have enabled robotic manipulators to perform increasingly complex manipulation tasks. These developments have the potential to bring robotic systems closer to deployment in dynamic and challenging disassembly scenarios. However, several limitations hinder their application to EOL product disassembly. IL frameworks are trained end-to-end and require large amounts of high-quality demonstrations to perform efficiently. Collecting such data for the disassembly of EOL products is difficult with current standard grippers, given the confined spaces and various asset configurations characteristic of EOL products. Similarly, existing RL algorithms for assembly and disassembly have solely been restricted to learning policies for relatively simple tasks with standard grippers, which limit their effectiveness in real-world EOL disassembly.

\begin{figure}[t]
    \centering
    \includegraphics[width=0.48\textwidth]{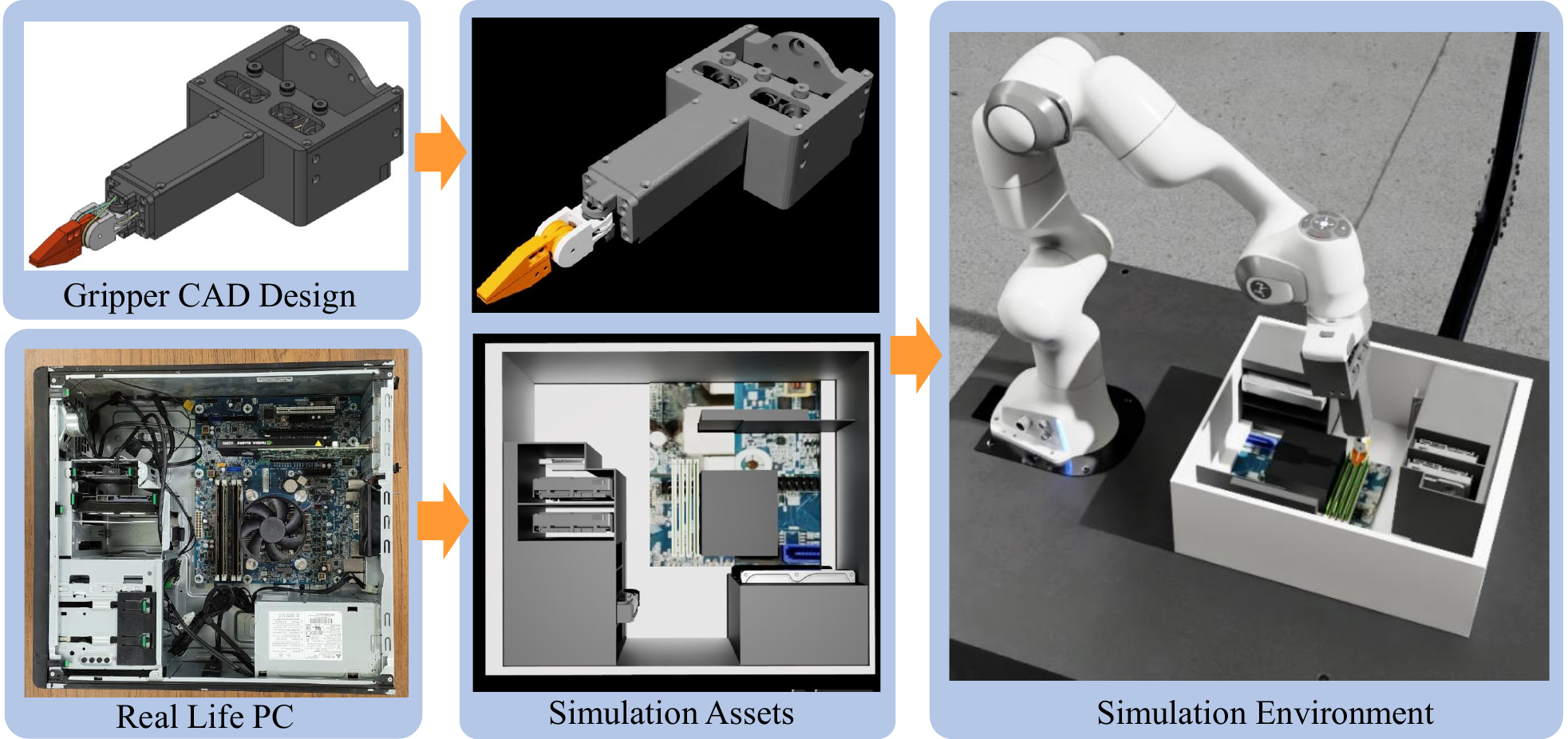}
    \caption{\grippername: A customized gripper for the disassembly of EOL desktops. We propose a customized gripper design and create simulation environments replicating EOL desktops. We then validate the effectiveness of the customized gripper within the simulation environment.}
    \label{fig:1_header_fig}
\end{figure}

EOL products disassembly requires complex manipulation policies that can adapt to components of varying sizes and configurations to establish greater automation in recycling. While some customized robotic systems have been developed for specific disassembly tasks \cite{rujanavech2016liam,lee2022robot,lee2024review,lee2022task,Liu2025}, to the best of our knowledge, no customized robotic gripper has been developed for the disassembly of EOL desktop components.

We introduce \grippername, a customized gripper designed for the disassembly of EOL computer desktops. Inspired by surgical robot grippers \cite{chandrasekaran2020design, sallam2023design, zhao2016estimating}, \grippername has 3 DOFs cable-driven joints, which enable the disassembly of desktop components with diverse configurations within confined spaces. In addition, we developed a physics-based disassembly simulation environment with a variety of tasks to test and showcase the effectiveness of \grippername for disassembly.

The contribution of this work is as follows:

\begin{itemize}
    \item We propose \grippername, a customized gripper for the disassembly of EOL desktops. When mounted on the end effector of robotic manipulators, the gripper provides a high level of dexterity, compared to conventional grippers for disassembly, particularly in confined spaces.
    \item We also develop a PC disassembly simulation environment in Isaac Sim, where \grippername is mounted on a robotic manipulator, and interacts with desktop assets of varying sizes and configurations. We demonstrate the effectiveness of our gripper using evaluations in this simulated environment across various tasks with different disassembly complexities.
\end{itemize}

This paper is organized as follows: Section \ref{sec:related-workd} reviews related work on intelligent robotic assembly and disassembly. Section \ref{sec: design} presents the design of \grippername. Section \ref{sec: simulation-env} describes the disassembly simulation environment and tasks, and Section \ref{sec: conclusion} concludes the paper.

%% file: Related-Work.tex
\section{Related Works} \label{sec:related-workd}
\noindent 
\textbf{Intelligent Robotic Assembly/Disassembly:} The use of robotic manipulators for assembly and disassembly tasks has been studied for decades. Traditional approaches have primarily utilized analytical methods, with a particular focus on peg-in-hole insertion task \cite{huang2013fast, tang2016autonomous, kimble2020benchmarking, von2020robots}. However, these methods are highly sensitive to modeling and sensing errors and struggle to generalize to complex or unseen environments. Recently, non-RL \cite{fu2022safe, spector2022insertionnet, wen2022you} and RL-based methods \cite{hou2020data, apolinarska2021robotic, luo2021learning, luo2021robust} have demonstrated high success rates, robustness to part geometry variations, and generalization across different part geometries. Despite these advances, current methods still rely on human demonstrations, initializations, and corrections, and require long training times to reach an acceptable level of success rate. 

High-fidelity physics simulators \cite{makoviychuk2021isaac, mittal2023orbit} have enabled sim-to-real methods for assembly and disassembly tasks, and offer parallelization that reduces training time \cite{noseworthy2025forge, tang2023industreal, guo2025srsa, tang2024automate}. However, these approaches primarily target peg-in-hole or NIST-style benchmark tasks \cite{davchev2022residual}, which differ significantly from the tasks involved in the disassembly of EOL products. Moreover, they typically employ conventional robotic grippers, that struggle to operate in the confined spaces of EOL product disassembly environments.

\begin{figure}[t]
    \centering
    \includegraphics[width=1\linewidth]{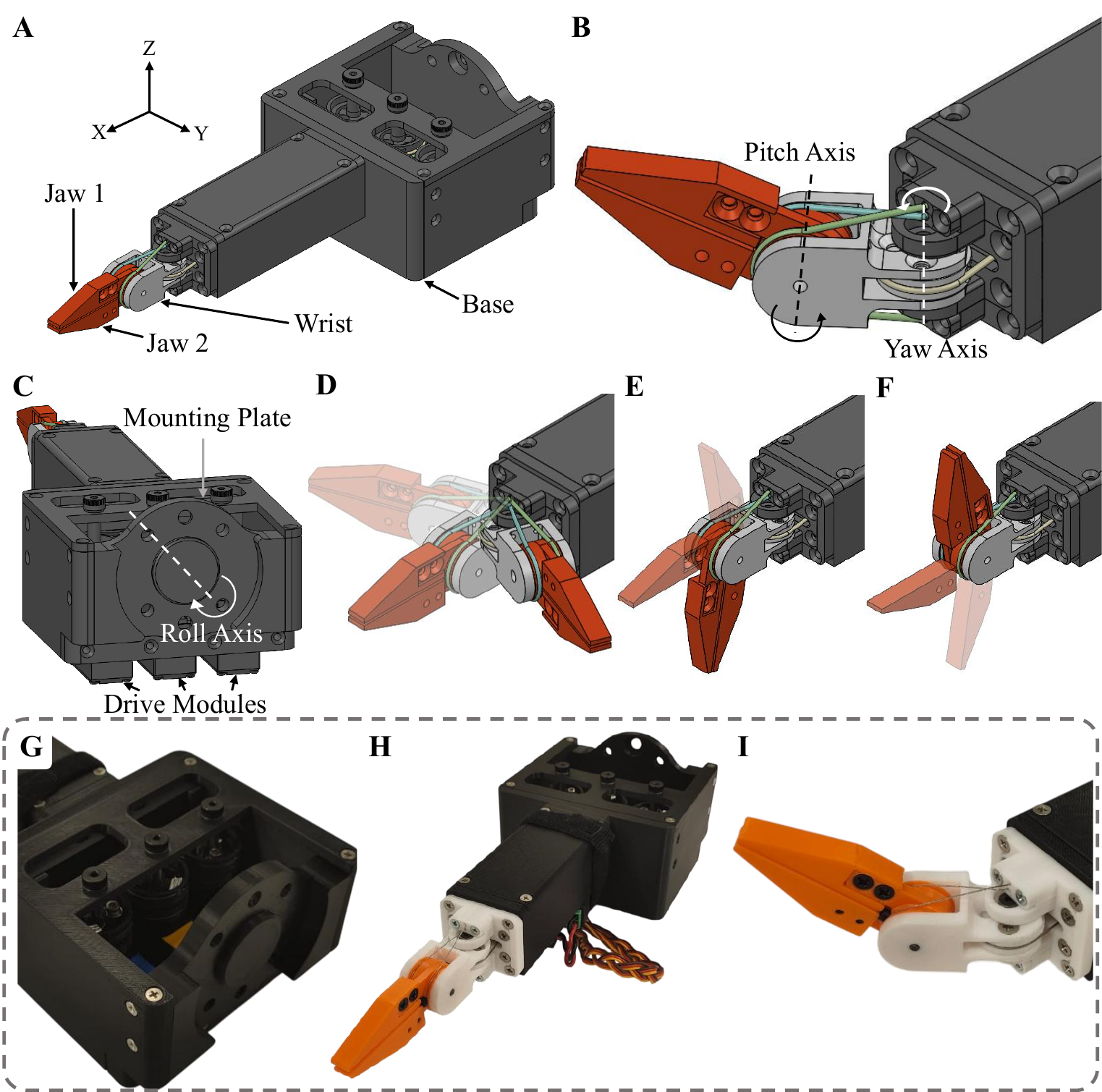}
    \caption{Overview of the \grippername design, with physical prototype. A. Gripper links: base, wrist, gripper 1, and gripper 2. B. Yaw and pitch DOFs of the gripper. C. The servo motors drive the joints, and the mounting plate attaches the gripper to robotic manipulators. D-F. Range of motion of gripper joints: D. wrist, E. jaw 1, and F. jaw 2. G-I. Photos of the gripper prototype.}
    \label{fig:2_gripper_design}
\end{figure}

\vspace{0.2cm}
\noindent
\textbf{Customized Robotic Grippers:} Conventional robotic grippers are widely used in robotic applications because of their simplicity and durability. Parallel jaw grippers such as Robotiq products \cite{robotiq_grippers_2025}, Franka Hand \cite{franka_hand_2025}, and the JACO gripper \cite{jain2016grasp} are capable of reliably grasp objects with precise and large forces \cite{weinberg2024survey}. But their limited dexterity and large size restrict their application in confined spaces characteristic of disassembly environments. Vacuum-driven grippers are capable of conforming to objects of arbitrary shapes \cite{brown2010universal}, yet their large size and limitations in grasping objects from arbitrary configurations impede their effectiveness for EOL product disassembly. Anthropomorphic hands \cite{shaw2023leap} provide high dexterity by mimicking the human hand, but their size and complexity limit their utilization in the disassembly of EOL products.

Recently, various domain-specific grippers have been developed to overcome the limitations of conventional grippers \cite{chandrasekaran2020design, sallam2023design, zhao2016estimating}. To achieve automation in disassembly, customized grippers have been developed for various disassembly tasks, such as removing snap-fit covers \cite{schumacher2013system}, unscrewing \cite{chen2014robot}, disassembling electric vehicle motors \cite{wegener2015robot}, dismantling LCD screens \cite{vongbunyong2015general, chen2020application}, and iPhones and their subcomponents \cite{rujanavech2016liam}.

For the disassembly of EOL desktop products, ElSayed \textit{et al.} \cite{elsayed2012robotic} proposed a system that integrates a robotic manipulator with sensing modalities for partial computer desktop disassembly. The KIT gripper \cite{klas2021kit} is a multifunctional tool designed for the disassembly of small electromechanical components and can be utilized for the smaller components within desktop computers. However, there is no customized gripper capable of disassembling computer desktops.

In this work, we present a customized gripper (\grippername) for the disassembly of EOL computer desktops. The gripper can access confined spaces within the disassembly environment and remove components in arbitrary configurations. We also provide a simulation platform that can be used for learning disassembly policies in simulation and transferring them to real-world for efficient desktop disassembly.

%% file: GripperDesign.tex
\section{Gripper Design} \label{sec: design}
This section presents the mechanical design and the actuation of the proposed gripper.

\begin{figure}[t]
    \centering
    \includegraphics[width=1\linewidth]{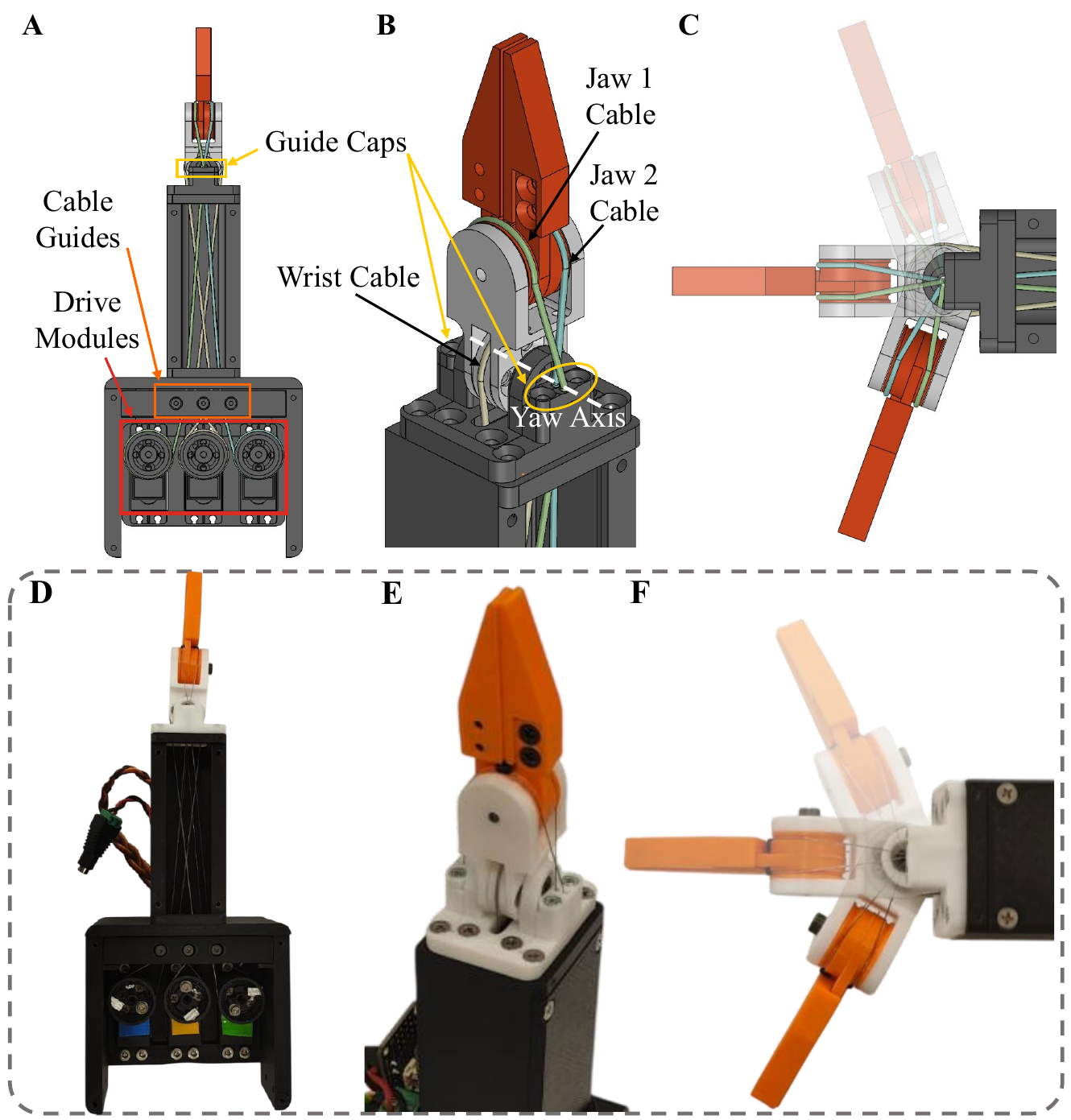}
    \caption{Cable transmission design of \grippername, with physical prototype. A. Cable-driven transmission mechanism. The cables are driven by the drive modules and pass through the cable guides. Cables for the jaws additionally pass through guide caps to establish a decoupled movement between joints. B-C. The length of the jaw cables remains the same as the wrist rotates, thanks to the guide caps on the wrist joint. D-F. Photos of the gripper prototype. D. gripper with covers removed, showing the cables. E. Close-up photo of the jaws. F. Cable length change is negligible as the wrist rotates.}
    \label{fig:3_cable_guiding}
\end{figure}

\subsection{Design Considerations}
There are two design considerations for the design of \grippername: \textbf{small size} and \textbf{high DOFs}.

\vspace{0.2cm}
\noindent
\textbf{Small Size}: The majority of commercial grippers are designed for open-space grasping and are too bulky for the confined spaces present within disassembly tasks. For example, the RAM chips on the PC motherboard are commonly spaced closer than 10 mm, and commercial two-finger grippers with their thick fingers struggle to grasp them without compromising and breaking other disassembly components.

\vspace{0.2cm}
\noindent
\textbf{High DOFs}: The internal structure of disassembly desktops varies significantly across vendors, resulting in large variations in the position and orientation of disassembly components.  These variations necessitate the gripper to operate in arbitrary positions and orientations for proper disassembly. Commercial grippers mainly rely on the movement of the supporting robotic manipulator to reach these arbitrary positions and orientations due to their low dexterity. This will lead the robotic manipulator to be in low manipulability, or even singularity regions, or even collide with other obstacles (e.g., human operator) in the disassembly environment.

With these design objectives in mind, we drew inspiration from surgical grippers \cite{sallam2023design} to design a disassembly gripper that features high dexterity and the ability to operate in confined spaces within the disassembly environment. The structure and design choices of the proposed gripper are demonstrated in \figref{fig:2_gripper_design}.

\subsection{Kinematics and Mechanical Design}
The kinematic structure of \grippername refers to the arrangement of its links and joints, which determine the poses and orientations it can achieve. It consists of four links: base, wrist, jaw 1, and jaw 2 (Figure \ref{fig:2_gripper_design}-A), and three joints (Figure \ref{fig:2_gripper_design}-B, D–F). Within this kinematic structure, the wrist can rotate w.r.t the base around the yaw axis, and the jaw can rotate w.r.t the wrist around the pitch axis. When mounted on a robotic manipulator, the additional degree of freedom (DOF) provided by the end effector allows the gripper to rotate around the roll axis, enabling full orientation control with a fixed end effector position (Figure \ref{fig:2_gripper_design}-C).

\begin{figure}[t]
    \centering
    \includegraphics[width=1\linewidth]{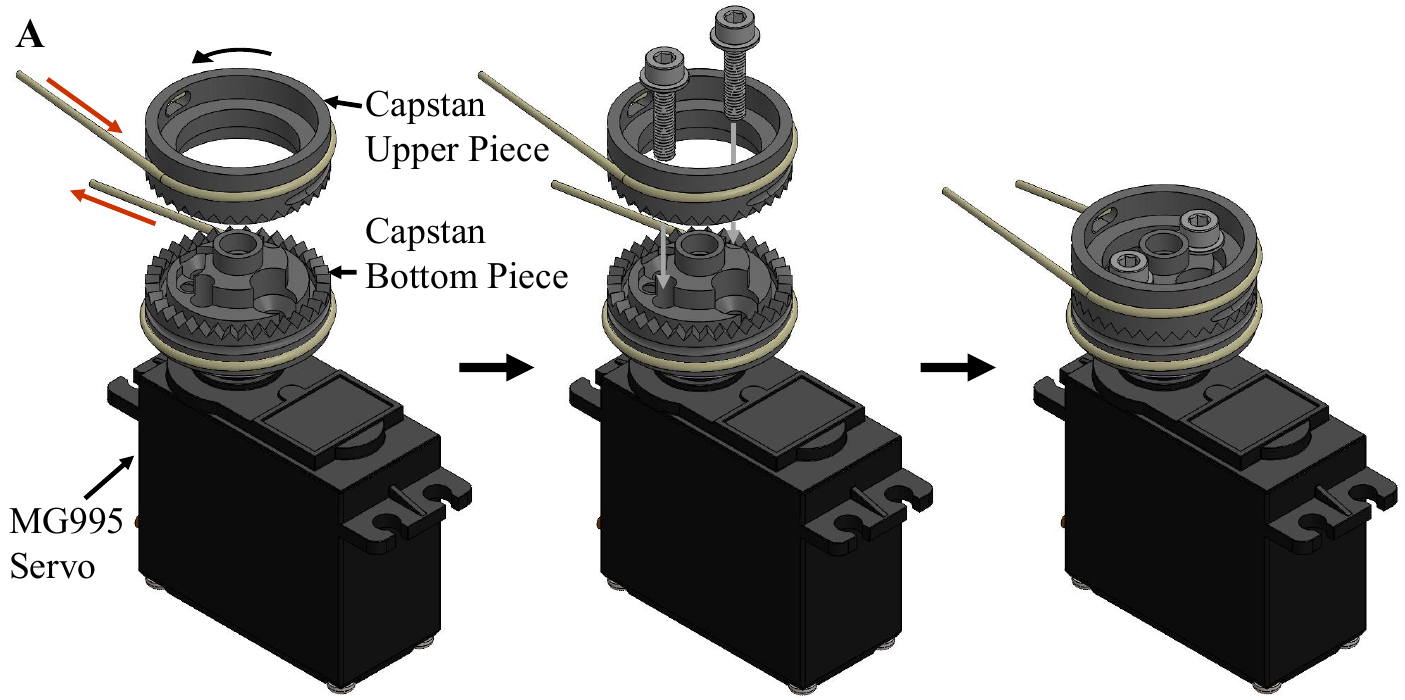}
    \caption{Tensioning process for the cable-driven transmission module of the \grippername. The capstan upper piece is first turned to pre-tension the cable. Then the pieces are secured with screws and locked in place by ratchets. This completes the drive module.}
    \label{fig:4_drive_module}
\end{figure}

\subsection{Actuation}
\grippername utilizes a cable-driven transmission mechanism \cite{chandrasekaran2020design} to actuate its joints, as shown in Figure \ref{fig:3_cable_guiding}. The \textbf{small size} design objective necessitates a transmission mechanism, since a direct-drive design would place actuators near the joints at the tip of the gripper, which is undesirable for disassembly in confined spaces. A more suitable design choice is to relocate the actuators away from the tip using a transmission mechanism. Linkage transmissions require additional moving links which increase the complexity of the gripper. Gears introduce friction and backlash, making them inefficient for this gripper. Belt-drive mechanisms have a non-negligible width, making them unsuitable at this scale. In contrast, cable-driven transmissions are lightweight and efficient, and do not require a fixed plane for transmission, which makes them well-suited for actuating joints across DOFs.

The cable routing of \grippername, as shown in \ref{fig:3_cable_guiding}, is designed as follows \cite{sallam2023design}: Each cable loop starts from each joints' respective drive module and passes through the cable guide to attach to the joint. After passing through the cable guide, the wrist cable directly attaches to the wrist joint, while cables for jaw joints are further passed through guide caps before being attached to the jaws.  The guide caps are designed such that their upper surface intersects the rotation axis (yaw axis) of the wrist. As a result, the cables bend at a sharp angle above the guiding cap surface rather than following a curve. When the wrist rotates, the cable length remains unchanged because the attachment point is on the rotation axis. This design choice decouples wrist rotation from jaw rotations, which simplifies the controller. Moreover, it enables joint force estimation from actuator currents \cite{zhao2016estimating}, without requiring actual force sensors on jaw links. This property is advantageous for our design, as adding force sensors to the gripper jaws would increase the overall size of the gripper.  Moreover, since the gripper interacts with rough EOL disassembly component surfaces, adding sensors to the jaws would complicate the design and make it less reliable.

The joints of the \grippername are driven with identical drive modules that feature a MG995 servo motor and a tensioning mechanism, as shown in Figure \ref{fig:3_cable_guiding}. Cable-driven mechanisms require pre-tensioning to prevent slacking effects. A common method is to use a pair of split capstans that can be rotated opposite to each other for pre-tensioning, then clamped to the drive shaft to maintain the tension. \cite{chandrasekaran2020design}. However, this approach requires customized metal parts which would complicate the manufacturing of the gripper. To keep the design compact, \grippername utilizes a pair of split capstans with ratchets that can lock onto each other when they are secured by screws axially. As shown in Figure \ref{fig:4_drive_module}, the capstan pieces can first be rotated to pre-tension the cables, and then secured together to complete the drive module. Thanks to this design choice, the design of \grippername remains compact, while demonstrating good strength with 3D-printed parts.

\begin{figure}
    \centering
    \includegraphics[width=.8\linewidth]{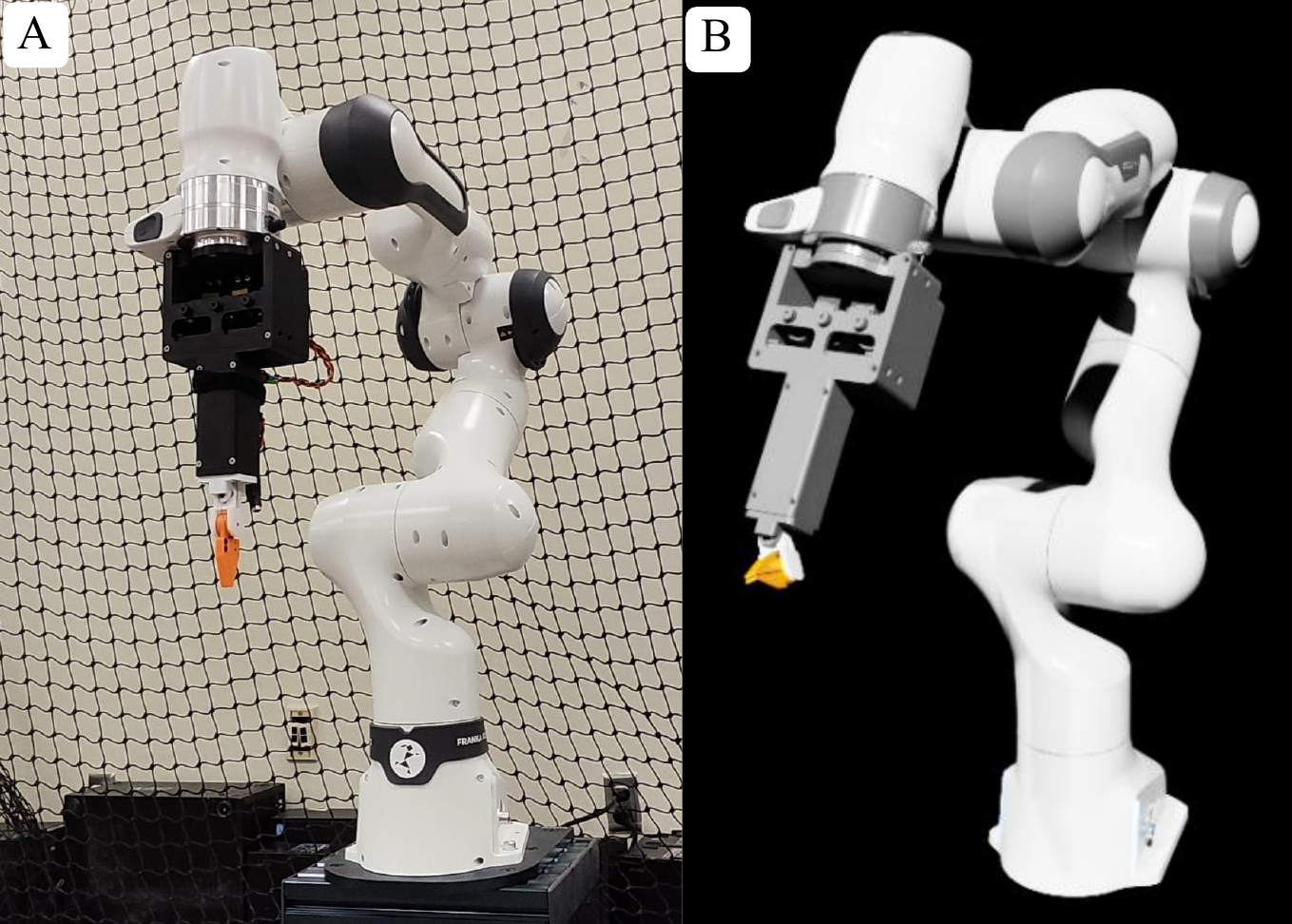}
    \caption{\grippername mounted on the robot, in reality and in simulation. A. Reality. B. Simulation.}
    \label{fig:5_robot_w_gripper}
\end{figure}

To verify that our design is manufacturable, we physically build a \grippername prototype and mount it on a Franka robotic arm (\figref{fig:5_robot_w_gripper}A). All the structural parts are 3D printed using Polylactic Acid (PLA), while other components are standard off-the-shelf parts. This demonstrates the feasibility of our design in the real world.

%% file: SimulationPlatform.tex
\section{\grippername Simulation Platform} \label{sec: simulation-env}
This section presents the validation of \grippername across various EOL computer desktop disassembly tasks in simulation.

\subsection{Simulation Environment and Disassembly Tasks}
\vspace{0.2cm}
\noindent
\textbf{\grippername:} To validate the effectiveness of \grippername in simulation, its CAD model is converted to a URDF and imported into the Isaac Sim environment \cite{NVIDIA_Isaac_Sim}. The gripper is mounted on the end effector of a Franka robotic arm (\figref{fig:5_robot_w_gripper}-B), and all \grippername joints are directly accessed and controlled within the environment.

\vspace{0.2cm}
\noindent
\textbf{Disassembly Tasks:} We would like to validate \grippername on several disassembly tasks with various complexities, which are common tasks within the disassembly of EOL computer desktops. The tasks include:

\begin{enumerate}
    \item Grasping and disassembling thin disassembly components.
    \item Grasping disassembly components located in confined spaces.
    \item Grasping disassembly components in arbitrary configurations (e.g., horizontal or vertical).
\end{enumerate}

\begin{figure}[t]
    \centering
    \includegraphics[width=1\linewidth]{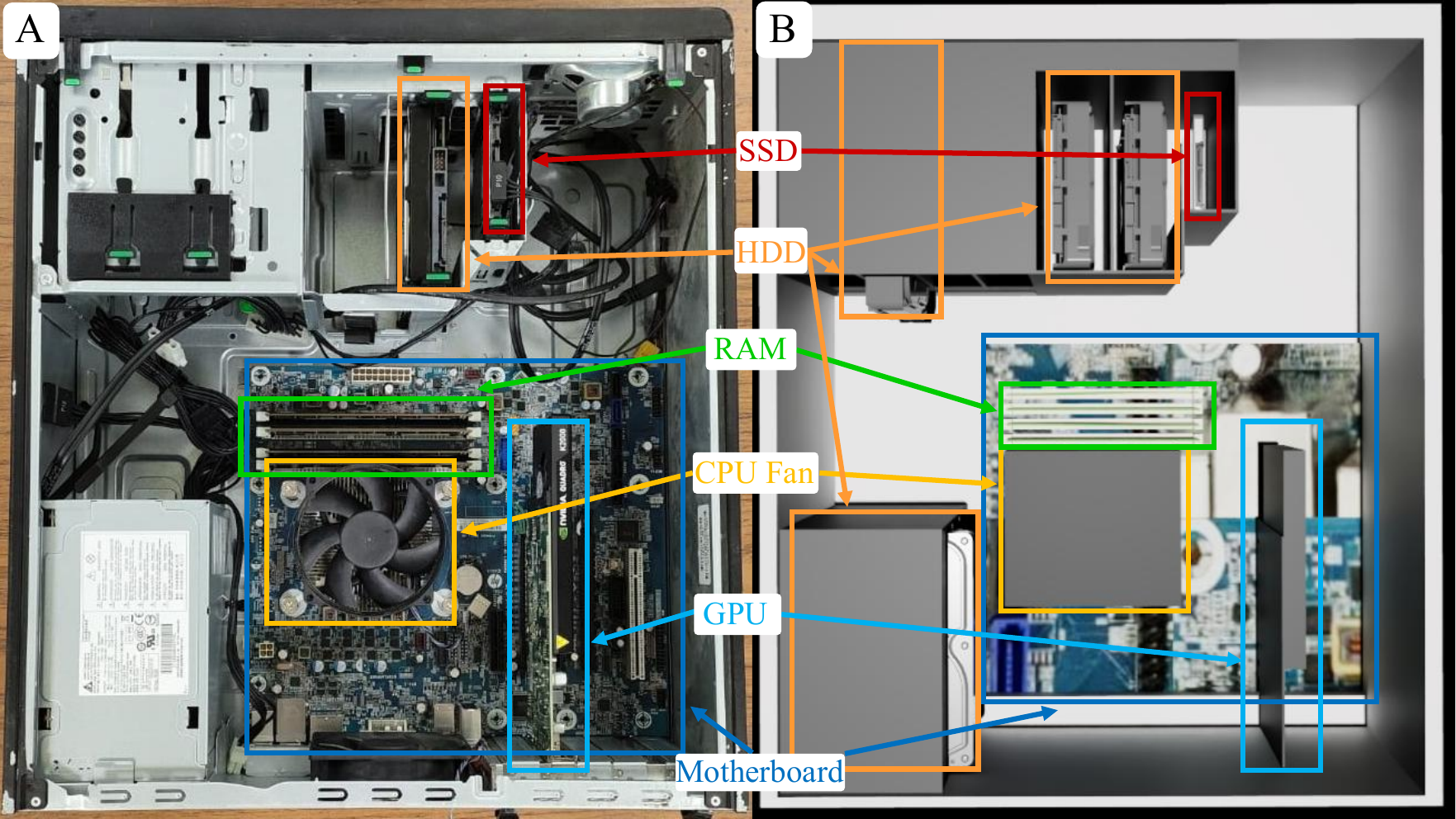}
    \caption{Real and simulated EOL desktops for \grippername evaluation. A. EOL HP Z230 Tower Workstation. B. Simplified simulated model of the EOL desktop within Isaac Sim. The horizontal and vertical HDDs are included in the simulation to represent variations across different desktops, although they are not present in the actual workstation shown.}
    \label{fig:6_pc_real2sim}
\end{figure}

\vspace{0.2cm}
\noindent
\textbf{Modeling of EOL Computer Desktops:} Based on the defined disassembly tasks, a computer desktop disassembly environment is constructed in the simulation environment. The model is derived from an EOL HP Z230 Tower Workstation (\figref{fig:6_pc_real2sim}-A), and is modeled using simplified CAD models. Most of the components serve as static obstacles, while the components to be disassembled are defined as free-moving rigid bodies that can interact with the robot and the environment. These include four Random-Access Memory (RAM) chips,  a Solid State Drive (SSD), and four Hard Disk Drives (HDDs). The form factor for SSDs is 2.5-inch, and the HDDs' form factor is 3.5-inch. For the purpose of the simulation, these specifications only affect their geometry. SSD and HDD models are obtained online \cite{grabcad_hdd} \cite{grabcad_ssd}.

Although the modeled PC has similar dimensions and layout to the real object, there are some key differences. For simplicity, cables are not modeled, as manipulating deformable objects remains a challenging task \cite{yin2021deformable}. Similarly, locking tabs and other mechanisms are not modeled, because they would either complicate the task or require the deployment of a second robotic manipulator. Finally, the horizontal and vertical HDDs shown in \figref{fig:6_pc_real2sim}-B are not present in the real setup. However, we include them because EOL PCs have various configurations, and we want our model to be representative of these possibilities.

The desktop has initial configurations in \figref{fig:6_pc_real2sim}-B and defined as follows. The RAM chips are slotted in their sockets and can be gripped using an upright pose. The SSD is slotted upright in a tight enclosure. Two HDDs are slotted upright near the SSD. The third one is slotted horizontally, while the last one is slotted vertically. The last two can not be gripped with an upright pose initially.

For disassembly task 1, we consider the grasping and removal of RAM chips within the modeled desktop. These chips are thin and closely spaced ($\approx$10mm in-between), making them difficult to manipulate with conventional grippers with wide and thick fingers. For disassembly task 2, we try to retrieve the SSD, which is slotted in a confined space that is difficult for regular grippers to reach into (width $\approx40\mathrm{mm}$). For disassembly task 3, we focus on HDDs. Their 3.5-inch form factor fit the grasp of most grippers, but they may be slotted in horizontal or vertical configurations. With conventional grippers, this requires the robot to re-orient its end effector using the entire arm, which often leads to configurations that would result in collision.

\begin{figure*}
    \centering
    \includegraphics[width=0.9\textwidth]{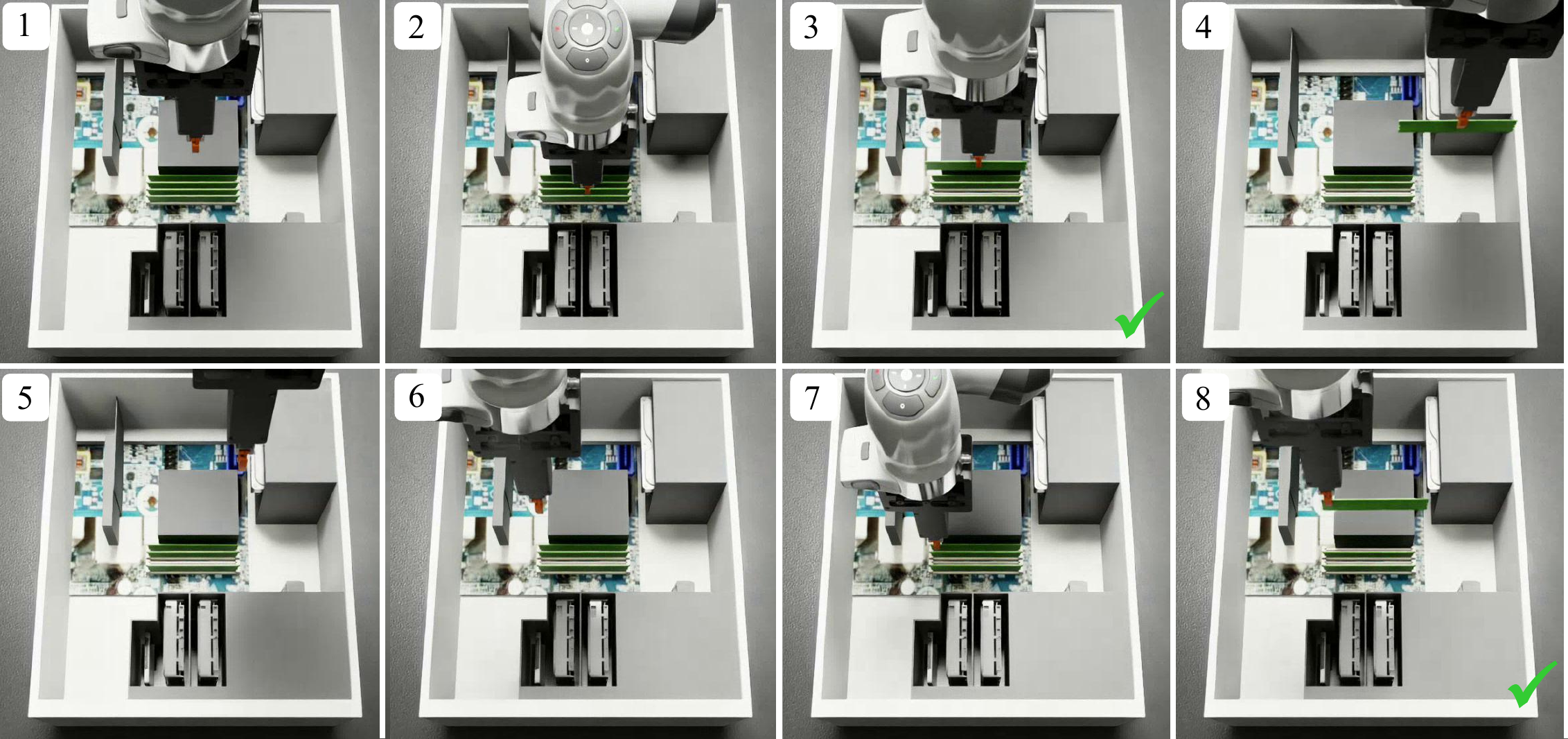}
    \caption{Task 1: thin component disassembly. The gripper is tasked with removing the third and first RAM chips (counting from up to bottom), which are the most confined and challenging of the RAM chips. 1-4: The disassembly sequence of the third RAM chip. 5-8: The disassembly sequence of the first RAM chip.}
    \label{fig:7_task1}
\end{figure*}

\begin{figure}
    \centering
    \includegraphics[width=1\linewidth]{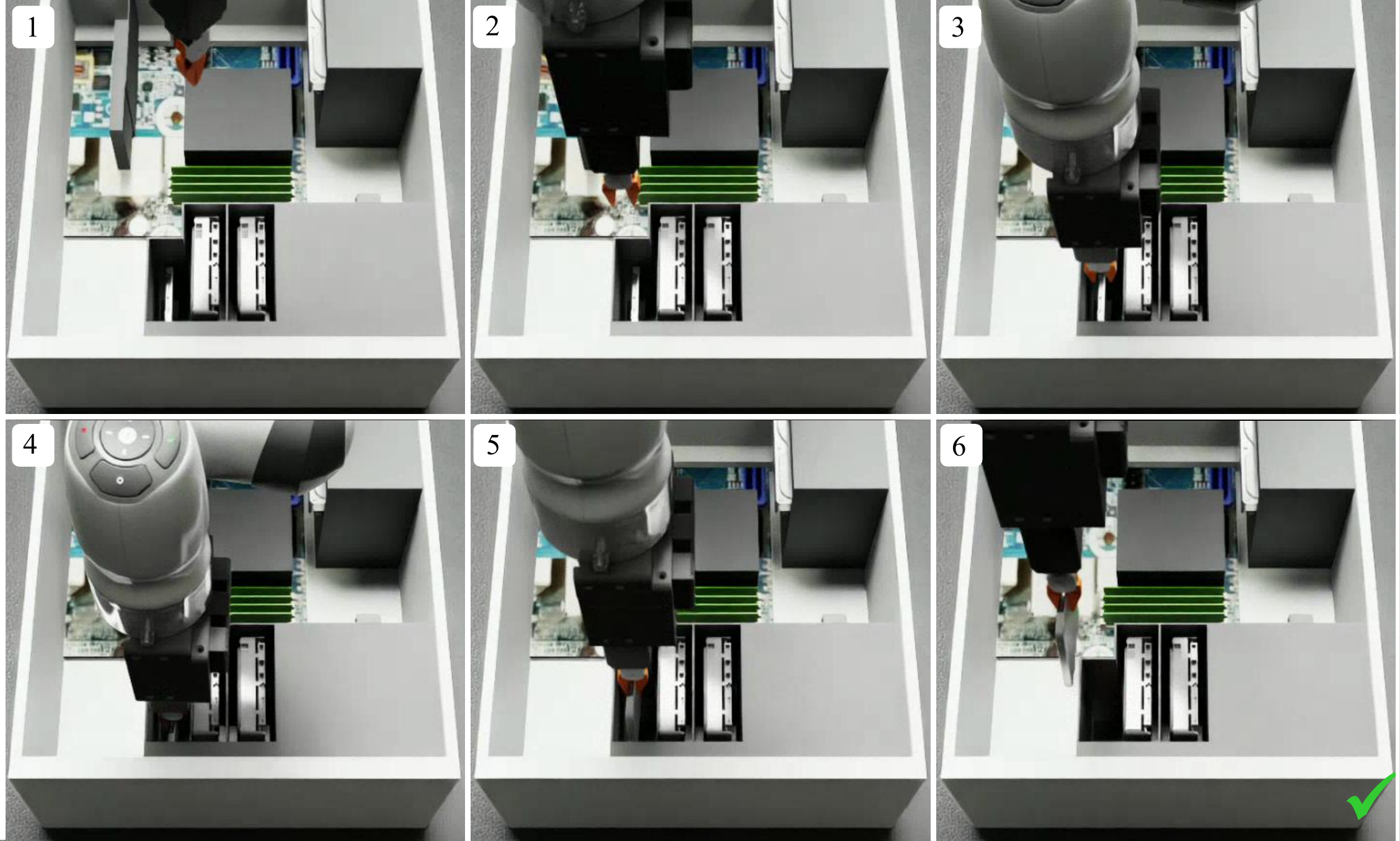}
    \caption{Task 2: disassembly within confined space. The gripper moves down the narrow enclosure and grasps the SSD and pulls it out.}
    \label{fig:8_task2}
\end{figure}

For the task, we disassemble both the horizontal and the vertical ones, divided into tasks 3a and 3b.

The simulation environment consists of the Franka robot with \grippername mounted on a table, with the EOL desktop in the front it, as shown in \figref{fig:1_header_fig}.

\begin{figure}
    \centering
    \includegraphics[width=0.5\textwidth]{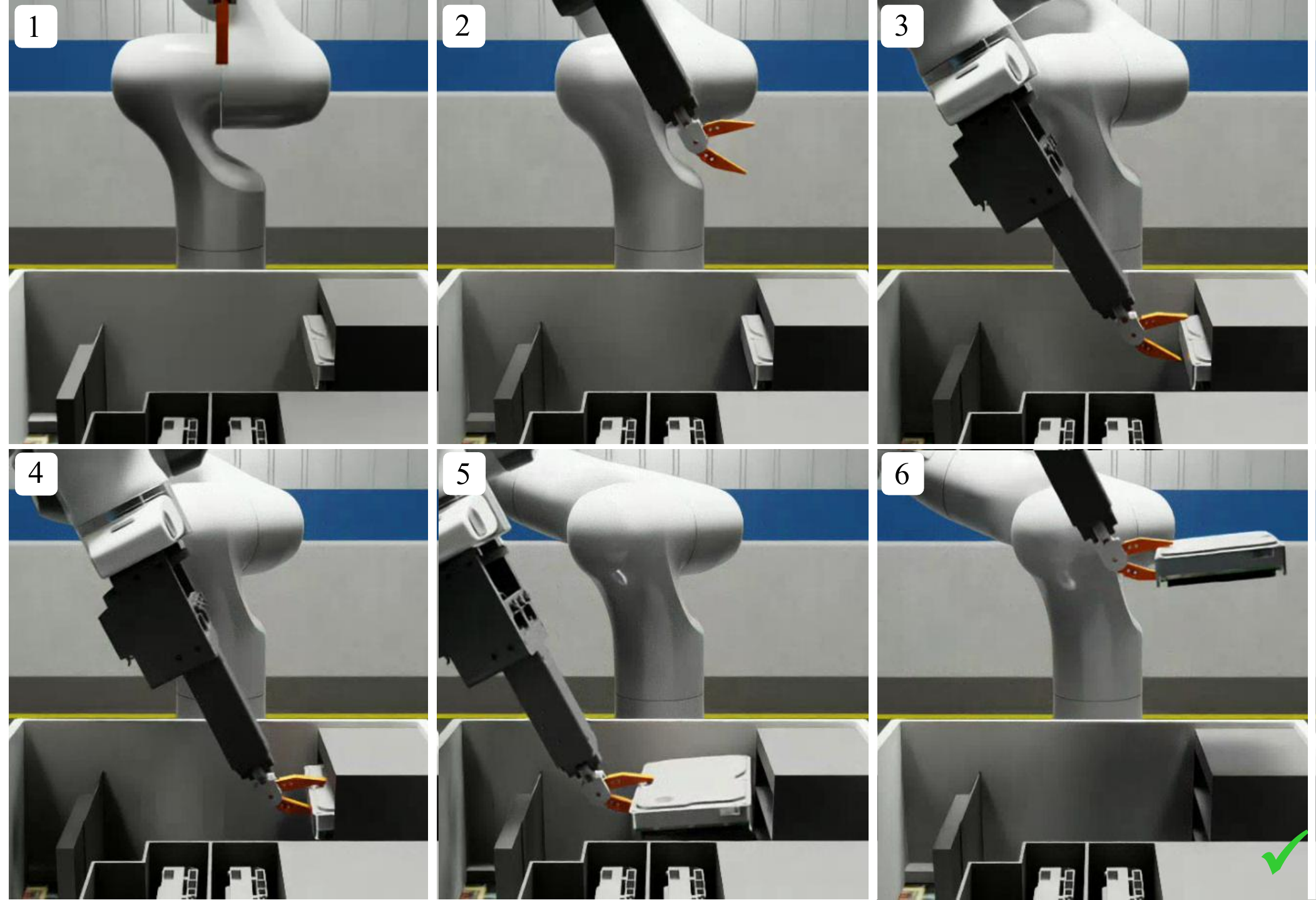}
    \caption{Task 3a: horizontal disassembly. The gripper successfully goes to a horizontal orientation, grasps the horizontal HDD, and pulls it out.}
    \label{fig:9_task3a}
\end{figure}

\subsection{Evaluations}
We conduct the evaluations of \grippername in the constructed simulation environment across the defined tasks.

\vspace{0.2cm}
\noindent
\textbf{Task 1:} This task demonstrates \grippername's capability to operate in confined spaces and disassemble thin components, as shown in Figure \ref{fig:7_task1}. The gripper is tasked to remove two of the most challenging RAM chips: The third (counting from up to bottom in \figref{fig:7_task1}), confined by neighboring chips on both sides, and the first, positioned very closely to the CPU fan, and accessible only through a side opening. As illustrated, the gripper utilizes its pointed jaws to fit between narrow gaps and successfully removes both chips. The remaining two chips are easier to remove and are therefore not included in this task. This experiment highlights \grippername’s capability to grasp thin, closely spaced objects.

\vspace{0.2cm}
\noindent
\textbf{Task 2:} This task demonstrates \grippername's ability to reach confined spaces as shown in \figref{fig:8_task2}. The gripper is tasked with disassembling the SSD, which requires accessing the enclosure. Since the gripper width is smaller than the enclosure opening, it can easily reach the SSD and remove it. This task highlights \grippername's effectiveness in confined space disassembly.

\begin{figure*}
    \centering
    \includegraphics[width=1\linewidth]{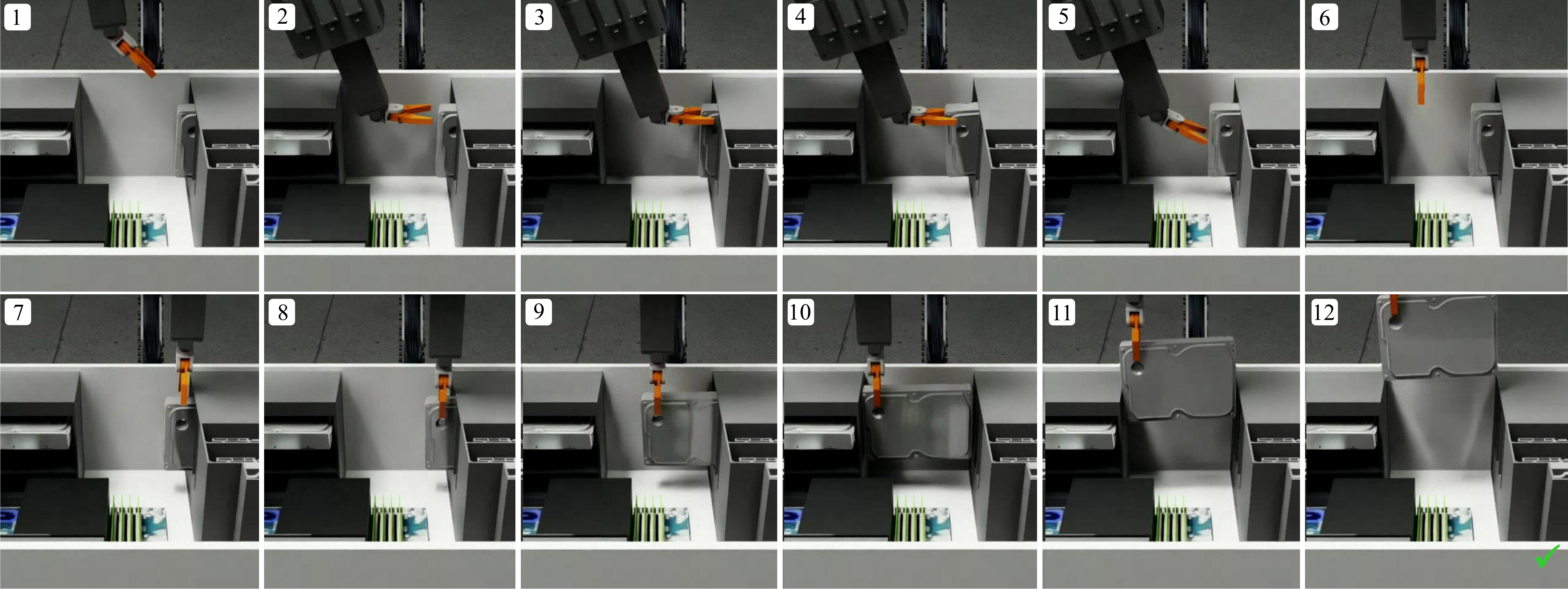}
    \caption{Task 3b: vertical disassembly. The gripper first partially pulls out the vertical HDD horizontally, then changes the grasp to fully pull it out in an upright pose due to the confined space of this task. 1. Re-orienting from the initial configuration. 2. Vertical grip orientation. 3. Gripping the HDD. 4. Pulling out the HDD. 5. Releasing HDD and reorienting. 6. Up-right grip Orientation. 7. Approaching HDD. 8. Gripping HDD. 9-10. pulling out HDD. 11-12. Successful disassembly.}
    \label{fig:10_task3b}
\end{figure*}

\vspace{0.2cm}
\noindent
\textbf{Task 3:} This task showcases the \grippername' ability to disassemble both vertical and horizontal components. Task 3a involves disassembling a horizontal HDD, as shown in \figref{fig:9_task3a}. The gripper reorients into a horizontal orientation, approaches the HDD from the side, grips it, pulls it out of the enclosure, and then lifts it out of the frame.

Task 3b involves disassembling an HDD in vertical orientation, as shown in \figref{fig:10_task3b}. The confined space of the EOL desktop prevents the gripper from pulling the HDD out horizontally. Instead, the gripper partially pulls out the HDD using a vertical grasp pose, switches to an upright pose, grips the newly exposed section, then pulls it out completely. This task demonstrates that our \grippername can handle disassembly components in both horizontal and vertical orientations, and through the continuity of the joint angles, generalize to arbitrary orientations. 

%% file: Conclusion.tex
\section{Conclusions and Future Works} \label{sec: conclusion}
In this paper, we introduced \grippername, a customized gripper for the disassembly of EOL computer desktops. \grippername features three degrees of freedom (DOF) and, when combined with the DOF of a robotic manipulator, can achieve arbitrary orientations within the disassembly environment. A cable-driven transmission mechanism was employed for the actuation of \grippername, which reduced its size and enabled operation in confined spaces. The cable-driven mechanism decouples the actuation of wrist and jaw joints, allowing jaw forces to be estimated directly from servo motor currents; a feature that is promising for learning disassembly policies.

To demonstrate \grippername’s effectiveness in operating within confined spaces and achieving arbitrary configurations (e.g., \grippername's design objectives), we developed an EOL desktop disassembly environment in Isaac Sim. The disassembly tasks were chosen to represent common operations across EOL desktops from different vendors. The evaluations demonstrated the effectiveness of \grippername for the disassembly of EOL desktops.

In future work, we plan to improve the hardware capabilities of \grippername and validate its effectiveness in physical disassembly tasks. In addition, we plan to use \grippername to learn disassembly policies for EOL desktops and move toward automated disassembly. Specifically, we will leverage the simulation environment developed in this paper to learn disassembly policies with reinforcement learning and transfer to real-world for effective disassembly. The proposed gripper can also be utilized to collect high-quality data for learning disassembly policies through imitation learning.